\documentclass[a4paper,english,graphicx,amscd,amsmath,amssymb,verbatim]{article}
\usepackage[T1]{fontenc}
\usepackage[latin9]{inputenc}
\usepackage{babel}

\usepackage{float}
\usepackage{amstext}
\usepackage{graphicx}
\usepackage[unicode=true, pdfusetitle,
 bookmarks=true,bookmarksnumbered=false,bookmarksopen=false,
 breaklinks=false,pdfborder={0 0 1},backref=false,colorlinks=false]
 {hyperref}

\begin{document}

\title{Adaptive Cluster Expansion (ACE): A Multilayer Network for Estimating
Probability Density Functions%
\thanks{This paper was submitted to IEEE Trans. NN on 16 December 1991. Paper
TNN \#1115. It was not accepted for publication, but it underpins
several subsequently published papers.%
}}

\author{S P Luttrell}

\maketitle
\noindent \textbf{Abstract:} We derive an adaptive hierarchical method
of estimating high dimensional probability density functions. We call
this method of density estimation the {}``adaptive cluster expansion'',
or ACE for short. We present an application of this approach, based
on a multilayer topographic mapping network, that adaptively estimates
the joint probability density function of the pixel values of an image,
and presents this result as a {}``probability image''. We apply
this to the problem of identifying statistically anomalous regions
in otherwise statistically homogeneous images.

\section{Introduction}

The purpose of this paper is to develop a novel type of adaptive network
for estimating probability density functions (PDF) for use in Bayesian
analysis \cite{Cox1946,Jeffreys1939}. We consider only techniques
that scale well for use in high dimensional spaces, such as the analysis
of large arrays of pixels in image processing. There are many attempts
to solve this type of density estimation problem. For instance, the
Boltzmann machine \cite{AckleyHintonSejnowski1985} is essentially
a trainable Gibbs distribution, which permits arbitrarily complicated
statistical structure to be modelled via hidden variables. Unfortunately,
this generality must be paid for by performing lengthy Monte Carlo
simulations. There are various extensions to this technique, such
as the higher order Boltzmann machine \cite{Sejnowski1986}, which
capture higher order statistical behaviour more economically, but
none of these variations has been shown to be suitable for high-dimensional
image processing problems. Using maximum entropy techniques \cite{Jaynes1982},
we develop a number of variations on the Gibbs distribution approach
\cite{Luttrell1989d}, and propose a scheme in which we replace simple
interactions between a large number of hidden variables (as in the
Boltzmann machine) by complicated interactions which directly model
the statistical structure of the data; this is an extreme form of
the approach taken in \cite{Sejnowski1986}.

The novel adaptive density estimator that we develop in \cite{Luttrell1989d}
is based on a multilayer network, in which we choose the layer-to-layer
connections to be hierarchical, and the layer-to-layer transformations
to be topographic mappings \cite{Kohonen1984}; this adaptively transforms
the input data into a multiscale {}``pyramid-like'' format. In \cite{Luttrell1989d}
we further propose that the joint PDFs of adjacent nodes in each layer
should be combined to form an estimate of the joint PDF of the nodes
in the input layer. By analogy with the standard derivation of Gibbs
distributions, we can also derive our joint PDF estimate by applying
the maximum entropy method \cite{Jaynes1982}. However, our result
is computationally much cheaper to implement than a standard Gibbs
distribution, because we do not need to perform Monte Carlo simulations
in order to integrate over the states of hidden variables. We suggest
the name {}``adaptive cluster expansion'' (ACE) for this type of
network estimate of high-dimensional joint PDFs. Other literature
on this approach can be found in \cite{Luttrell1989c,Luttrell1989b,Luttrell1989e,Luttrell1990b,Luttrell1991a},
where we further develop multilayer topographic mapping networks,
and their relationship to vector quantisers.

The purpose of this paper is to present a complete account of ACE,
and to demonstrate its effectiveness when applied to the problem of
density estimation. We do not dwell on the details of how to implement
the topographic mapping training algorithm (we review this in the
appendix). In Section \ref{XRef-Section-35153624} we develop the
ACE method of density estimation by appealing to simple counting arguments.
In Section \ref{XRef-Section-35153635} we demonstrate the power of
ACE by applying it to the problem of estimating the joint PDF of the
pixels of textured images selected from the Brodatz album \cite{Brodatz1966}.

\section{Probability Density Function Estimation}

\label{XRef-Section-35153624}

In this section we present a derivation of the ACE estimate $Q(\text{\boldmath\ensuremath{x}})$
of a PDF $P(\text{\boldmath\ensuremath{x}})$. We develop this result
by appealing to simple counting arguments and by using a diagrammatic
language.

\subsection{Derivation of the ACE Estimate of a PDF}

\begin{figure}[h]

\begin{centering}
\includegraphics[width=7cm]{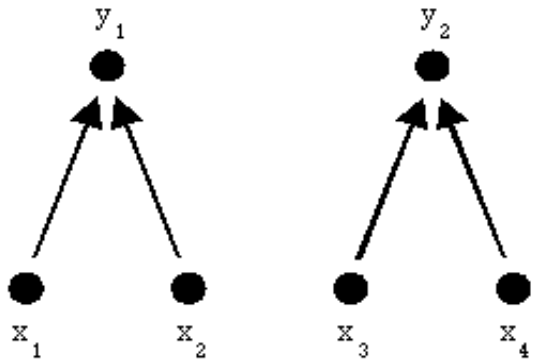}
\par\end{centering}

\caption{Basic 2-layer network. The input space (layer 0) is 4-dimensional
and the output space (layer 1) is 2-dimensional. The layer 0-to-1
transformation factorises into two independent transformations: $y_{1}$
depends only on $(x_{1},x_{2})$, and $y_{2}$ depends only on $(x_{3},x_{4})$.}

\label{XRef-Figure-3515426}
\end{figure}

We show a simple network in Figure \ref{XRef-Figure-3515426}, where
the input space is 4-dimensional, the output space is 2-dimensional,
and we factorise the feedforward transformation as $\text{\boldmath\ensuremath{y}}(\text{\boldmath\ensuremath{x}})=(y_{1}(x_{1},x_{2}),y_{2}(x_{3},x_{4}))$.
Suppose we estimate the joint PDF $P_{\mathrm{out}}(y_{1},y_{2})$
of the outputs, and the joint PDFs $P_{\mathrm{in},12}(x_{1},x_{2})$
and $P_{\mathrm{in},34}(x_{3},x_{4})$ of each pair of inputs, by
measuring their histograms, for instance. Using this information alone
we now wish to construct an estimate $Q(\text{\boldmath\ensuremath{x}})$
of the true joint PDF $P(\text{\boldmath\ensuremath{x}})$ of the
4-dimensional input. There are two alternative, but equivalent, ways
of writing $Q(\text{\boldmath\ensuremath{x}})$, each of which has
its own interesting interpretation.

Firstly, we may write

\begin{equation}
Q(\text{\boldmath\ensuremath{x}})=P_{\mathrm{out}}(y_{1}(x_{1},x_{2}),y_{2}(x_{3},x_{4}))\,\frac{P_{\mathrm{in},12}(x_{1},x_{2})}{P_{\mathrm{out}}(y_{1}(x_{1},x_{2}))}\,\frac{P_{\mathrm{in},34}(x_{3},x_{4})}{P_{\mathrm{out}}(y_{2}(x_{3},x_{4}))}\label{XRef-Equation-35154256}\end{equation}

\noindent In Equation \ref{XRef-Equation-35154256} we construct $Q(\text{\boldmath\ensuremath{x}})$
as follows. We use $P_{\mathrm{out}}(y_{1},y_{2})$ directly to estimate
the joint PDF of the outputs, and indirectly to estimate the joint
PDF of the inputs. In order to convert a PDF in output space (i.e.
$P_{\mathrm{out}}(y_{1},y_{2})$) into a PDF in input space we must
divide $P_{\mathrm{out}}(y_{1},y_{2})$ by a compression factor equal
to the number of input values that can produce the observed output
value. Because we obtain $y_{1}$ and $y_{2}$ \textit{separately}
from the pairs $(x_{1},x_{2})$ and $(x_{3},x_{4})$, respectively,
this compression factor is the product of two separate factors. For
instance, the compression factor corresponding to $y_{1}$ is the
ratio $\frac{P_{\mathrm{out}}(y_{1}(x_{1},x_{2}))}{\langle P_{\mathrm{in},12}(x_{1},x_{2})\rangle}$,
where $\langle P_{\mathrm{in},12}(x_{1},x_{2})\rangle$ is the average
value of $P_{\mathrm{in},12}(x_{1},x_{2})$ over all the $(x_{1},x_{2})$
that produce the same value of $y_{1}$. However, we may refine this
compression factor by using $P_{\mathrm{in},12}(x_{1},x_{2})$ instead
of $\langle P_{\mathrm{in},12}(x_{1},x_{2})\rangle$ in the denominator,
to yield the ratio $\frac{P_{\mathrm{out}}(y_{1}(x_{1},x_{2}))}{P_{\mathrm{in},12}(x_{1},x_{2})}$.
An analogous argument may be applied to obtain the compression factor
corresponding to $y_{2}$, and the results combined to obtain the
final expression for $Q(\text{\boldmath\ensuremath{x}})$ as shown
in Equation \ref{XRef-Equation-35154256}. \begin{equation}
Q(\text{\boldmath\ensuremath{x}})=P_{\mathrm{in},12}(x_{1},x_{2})P_{\mathrm{in},34}(x_{3},x_{4})\,\frac{P_{\mathrm{out}}(y_{1}(x_{1},x_{2}),y_{2}(x_{3},x_{4}))}{P_{\mathrm{out}}(y_{1}(x_{1},x_{2}))\, P_{\mathrm{out}}(y_{2}(x_{3},x_{4}))}\label{XRef-Equation-35154416}\end{equation}

\noindent which is trivially the same as Equation \ref{XRef-Equation-35154256},
but we have arranged its terms in a new way. This furnishes us with
an alternative interpretation of $Q(\text{\boldmath\ensuremath{x}})$.
Thus, imagine that we are provided only with $P_{\mathrm{in},12}(x_{1},x_{2})$
and $P_{\mathrm{in},34}(x_{3},x_{4})$, and no information about the
correlations between the pair $(x_{1},x_{2})$ and the pair $(x_{3},x_{4})$.
This is sufficient for us to construct $Q(\text{\boldmath\ensuremath{x}})$
as the product $P_{\mathrm{in},12}(x_{1},x_{2})\, P_{\mathrm{in},34}(x_{3},x_{4})$.
Now, we admit that in fact we also know $P_{\mathrm{out}}(y_{1},y_{2})$,
which is a source of information about correlations between the pair
$(x_{1},x_{2})$ and the pair $(x_{3},x_{4})$. We make use of this
information by forming the dimensionless ratio $\frac{P_{\mathrm{out}}(y_{1},y_{2})}{P_{\mathrm{out}}(y_{1})\, P_{\mathrm{out}}(y_{2})}$,
which differs from unity when $y_{1}$ and $y_{2}$ are correlated
random variables (i.e. $P_{\mathrm{out}}(y_{1},y_{2})\neq P_{\mathrm{out}}(y_{1})\, P_{\mathrm{out}}(y_{2})$).
This ratio is greater (or less) than unity when the pair $(y_{1},y_{2})$
is more (or less) likely to occur than would have been estimated from
knowledge of the marginal PDFs $P_{\mathrm{out}}(y_{1})$ and $P_{\mathrm{out}}(y_{2})$
alone. Finally, we use this dimensionless ratio as a correction factor
to obtain the expression for $Q(\text{\boldmath\ensuremath{x}})$
shown in Equation \ref{XRef-Equation-35154416}. This derivation is
heuristic, but it leads to the same result as shown in Equation \ref{XRef-Equation-35154256}.
\begin{figure}[h]

\begin{centering}
\includegraphics[width=7cm]{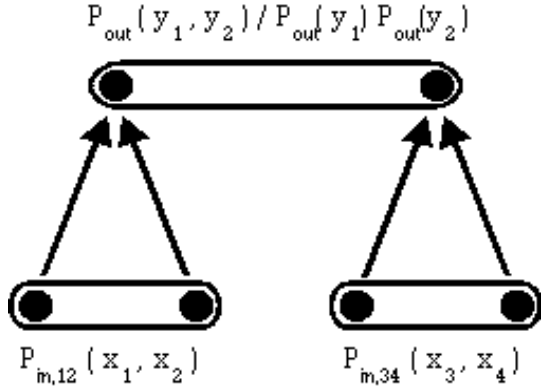}
\par\end{centering}

\caption{The clique PDFs that we use in the basic 2-layer network. $P_{\mathrm{out}}(y_{1},y_{2})$
is the joint PDF of the pair of network outputs, and $P_{\mathrm{out}}(y_{1})$
and $P_{\mathrm{out}}(y_{2})$ are its two marginal PDFs. $\frac{P_{\mathrm{out}}(y_{1},y_{2})}{P_{\mathrm{out}}(y_{1})\, P_{\mathrm{out}}(y_{2})}$
is a dimensionless ratio which records correlations between $y_{1}$
and $y_{2}$. $P_{\mathrm{in},12}(x_{1},x_{2})$ and $P_{\mathrm{in},34}(x_{3},x_{4})$
are the joint PDFs of the pairs of inputs from which $y_{1}$ and
$y_{2}$ derive, respectively.}

\label{XRef-Figure-35154532}
\end{figure}

In Figure \ref{XRef-Figure-35154532} we present an alternative representation
of the network in Figure \ref{XRef-Figure-3515426}, in which we emphasise
the PDFs that we use to construct $Q(\text{\boldmath\ensuremath{x}})$.
Thus we introduce a shorthand notation in which we use an oval to
highlight each clique of nodes in the network. We define the word
{}``clique'' to mean {}``complete set of nodes having the same
parent node''. As is conventional when discussing tree-like networks,
we regard the higher layers of the network as being the ancestors
of the lower layers, regardless of the fact that the direction of
information flow is in the opposite direction through the tree. We
then construct $Q(\text{\boldmath\ensuremath{x}})$ as the product
of the three clique PDFs shown, whilst ensuring that the clique in
layer 1 is appropriately normalised to render its contribution dimensionless.
This leads to the form of $Q(\text{\boldmath\ensuremath{x}})$ in
Equation \ref{XRef-Equation-35154416}.

This diagrammatic approach to constructing $Q(\text{\boldmath\ensuremath{x}})$
may be readily extended to any tree-like feedforward network. We favour
this approach, because the basic strategy for deriving $Q(\text{\boldmath\ensuremath{x}})$
by invoking compression factors remains the same, but the burden of
notational detail becomes somewhat heavy, so diagrams provide an ideal
shortcut. For convenience, we summarise the prescription for constructing
$Q(\text{\boldmath\ensuremath{x}})$ from a tree-like diagram as follows:
\begin{enumerate}
\item Estimate all of the clique PDFs, as histograms, for instance.
\item Deduce all of the single-node marginal PDFs from the clique PDFs estimated
in the previous step. For instance this would create $P_{\mathrm{out}}(y_{1})$
and $P_{\mathrm{out}}(y_{2})$ from $P_{\mathrm{out}}(y_{1},y_{2})$.
This step is not needed in in layer 0.
\item From the results estimated in the previous two steps, for each clique
compute a clique factor as follows:\label{Step:3}

\begin{enumerate}
\item In the input layer the factor is the clique PDF itself.
\item In other layers the factor is the clique PDF divided by the product
of its marginal PDFs (e.g. $\frac{P_{\mathrm{out}}(y_{1},y_{2})}{P_{\mathrm{out}}(y_{1})\, P_{\mathrm{out}}(y_{2})}$).
\end{enumerate}
\item Finally, to construct $Q(\text{\boldmath\ensuremath{x}})$, form the
product of all of the clique factors estimated in the previous step.
\end{enumerate}

\subsection{Translation Invariance}

A disadvantage of the above prescription for constructing $Q(\text{\boldmath\ensuremath{x}})$
is that it does not treat the components of $\text{\boldmath\ensuremath{x}}$
on an equal footing. For instance, in Equation \ref{XRef-Equation-35154416}
we see that the pair $(x_{1},x_{2})$ is treated differently from
the pair $(x_{2},x_{3})$, even though both of these are pairs of
adjacent components in the data. In order to solve this problem we
construct a number of different tree-like networks, each of which
breaks symmetry in its own peculiar way, and then we combine the results
from each network to construct a composite $Q(\text{\boldmath\ensuremath{x}})$
which respects the required symmetry. %
\begin{figure}[h]

\begin{centering}
\includegraphics[width=7cm]{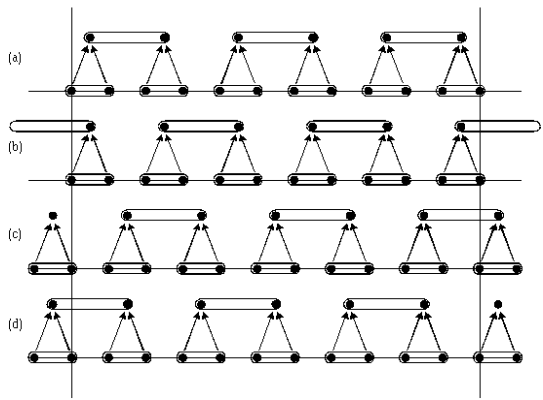}
\par\end{centering}

\caption{An example of the 4 separate 2-layer networks that we need to combine
in order to produce a $Q(\text{\boldmath\ensuremath{x}})$ that treats
each component of $\text{\boldmath\ensuremath{x}}$ on an equal footing.
Figure 3a shows the basic 2-layer network, Figure 3b shows the same
network with the layer 1 clique PDFs translated. Figure 3c and Figure
3d derive from Figure 3a and Figure 3b by simultaneously translating
the clique PDFs in both network layers.}

\label{XRef-Figure-35154955}
\end{figure}

\noindent In Figure \ref{XRef-Figure-35154955} we show an example
of the set of 4 different 2-layer networks which we need to combine
in order to construct a composite $Q(\text{\boldmath\ensuremath{x}})$.
In this example we assume that the input is a high dimensional vector,
so we can ignore edge effects. We replicate the basic network structure
of Figure \ref{XRef-Figure-35154532} across the input vector, as
shown. Each of the 4 networks has its own set of clique PDFs (drawn
as ovals in Figure \ref{XRef-Figure-35154532}), each of which leads
to its own estimate $Q(\text{\boldmath\ensuremath{x}})$ which breaks
symmetry. However, a symmetric combination (such as the arithmetic
or geometric mean) of these 4 results treats each component of $\text{\boldmath\ensuremath{x}}$
on an equal footing. We can verify this by noting that the set of
cliques that contributes in the spatial neighbourhood of each component
of $\text{\boldmath\ensuremath{x}}$ does not depend (apart from a
trivial overall translation) on which component we select.

We must select a prescription for forming the composite $Q(\text{\boldmath\ensuremath{x}})$.
It needs only to be a symmetric combination of the 4 individual estimates
that we show in Figure \ref{XRef-Figure-35154955}; the arithmetic
mean and geometric mean are obvious choices. On pragmatic grounds,
we choose to use the geometric mean, because it corresponds to the
arithmetic mean of $\log Q(\text{\boldmath\ensuremath{x}})$, which
is more convenient to perform in limited precision hardware ($\log Q(\text{\boldmath\ensuremath{x}})$
has a much smaller dynamic range than $Q(\text{\boldmath\ensuremath{x}})$,
assuming that we avoid the logarithmic singularity). %
\begin{figure}[h]

\begin{centering}
\includegraphics[width=7cm]{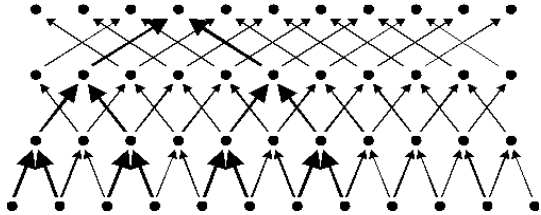}
\par\end{centering}

\caption{Example of the composite network connectivity that we require in order
for a single network to compute a composite $Q(\text{\boldmath\ensuremath{x}})$,
which treats each component of the input on an equal footing. This
connectivity is the union of all of the binary trees can be generated
from a reference binary tree (which we highlight in bold).}

\label{XRef-Figure-35155131}
\end{figure}

In Figure \ref{XRef-Figure-35155131} we show the connectivity of
part of a 4-layer composite network that can be used to process the
input data in preparation for constructing a composite $Q(\text{\boldmath\ensuremath{x}})$.
This connectivity contains all possible embedded tree-like networks,
and in Figure \ref{XRef-Figure-35155131} we highlight one such embedded
tree for illustrative purposes.

For an $n$-layer network, we form the composite $Q(\text{\boldmath\ensuremath{x}})$
as the geometric mean over the $Q(\text{\boldmath\ensuremath{x}})$
derived from all tree-like networks that are embedded in this composite
network, to yield the geometric mean PDF $Q_{\mathrm{gm}}(\text{\boldmath\ensuremath{x}})$
in the form \begin{equation}
\log Q_{\mathrm{gm}}(\text{\boldmath\ensuremath{x}})=\sum\limits _{L=0}^{n-1}\frac{1}{2^{L+1}}\sum\limits _{k}\log P_{k}^{L}\label{XRef-Equation-35155241}\end{equation}

\noindent where $L$ sums over layers 0 to $n-1$ of the network,
$k$ sums over cliques within a layer of the network, and $P_{k}^{L}$
is the clique PDF at position $k$ in layer $L$. It is important
to note that the cliques are not simply adjacent nodes in each layer
of the network. We must select pairs of nodes that form a {}``complete
set of nodes having the same parent node''. In layer 0 this means
that the nodes are adjacent. In layer 1 the nodes in a pair are separated
by 1 intervening node. In layer 2 there are 3 intervening nodes, and
so on. For $L\geq1$ we must ensure that the $P_{k}^{L}$ are dimensionless
by dividing out the marginal PDFs, as in Equation \ref{XRef-Equation-35154416}.
The $\frac{1}{2^{L+1}}$ factor ensures that we include each tree-like
network exactly once, and that the final result is indeed the geometric
mean of these contributions. Figure \ref{XRef-Figure-35154955} shows
the terms that Equation \ref{XRef-Equation-35155241} generates when
we set $n=2$.

There are two further assumptions that we could make in order to simplify
our result even further. Firstly, we could assume that the layer-to-layer
transformations in Figure \ref{XRef-Figure-35155131} were independent
of position $k$ within each layer $L$. Secondly, we could assume
that the clique PDFs were independent of position $k$ within each
layer $L$. We can make both of these assumptions if the statistical
properties of the input data are known to be translationally invariant
(such as might be the case for an image of a texture, for instance).
In all of our numerical simulations we make these two simplifying
assumptions.

\subsection{Modular Implementation}

We now describe a practical implemention of Equation \ref{XRef-Equation-35155241}
in the context of image processing (i.e. 2-dimensional arrays of pixels
of data). There are three basic operations to perform. We must use
a training set to determine suitable layer-to-layer transformations,
then estimate the clique PDFs in each network layer, and then construct
$\log Q_{\mathrm{gm}}(\text{\boldmath\ensuremath{x}})$ from these
estimates. Ideally we should optimise the layer-to-layer transformations
directly so that the constructed $Q_{\mathrm{gm}}(\text{\boldmath\ensuremath{x}})$
is {}``close to'' $P(\text{\boldmath\ensuremath{x}})$ in some sense
(e.g. relative entropy), but we have not yet found a computationally
cheap way of doing this. Instead, we tackle the problem indirectly,
by using our existing multilayer topographic mapping network technique
\cite{Luttrell1989e}. There are two main reasons for this choice.
Firstly, this type of network is computationally cheap to train; we
typically train such a network at the rate of 2.3 second per layer
on a VAXstation 3100 workstation (assuming 6 bit data values). Secondly,
the network encodes the input in such a way as to be able to reconstruct
it approximately from the state of any network layer. Although this
second property does not in general imply that the encoded input is
the optimal one for constructing an estimate of the input PDF, it
turns out that it does produce useful results. %
\begin{figure}[h]

\begin{centering}
\includegraphics[width=7cm]{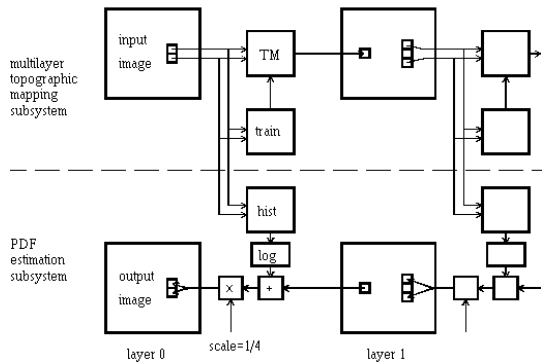}
\par\end{centering}

\caption{First two layers of a modular system for constructing $Q_{\mathrm{gm}}(\text{\boldmath\ensuremath{x}})$.
The top half of the diagram is a multilayer network subsystem (actually,
a multilayer topographic mapping in this case), which operates from
left to right. The bottom half of the diagram is the PDF estimation
subsystem, which operates from right to left. We connect the two systems
by feeding logarithmic clique PDFs measured in the multilayer network
through to be added together in the PDF estimator.}

\label{XRef-Figure-35155514}
\end{figure}

In Figure \ref{XRef-Figure-35155514} we show a system for constructing
$Q_{\mathrm{gm}}(\text{\boldmath\ensuremath{x}})$, which consists
of two interconnected subsystems - a multilayer topographic mapping
subsystem for transforming the input image, and a PDF estimation subsystem
for forming an output image which contains the contributions to $Q_{\mathrm{gm}}(\text{\boldmath\ensuremath{x}})$,
each recorded in its spatially correct location in the image. For
obvious reasons, we call the output image a {}``probability image''.
The flow from left to right across the top half of Figure \ref{XRef-Figure-35155514}
implements the network structure in Figure \ref{XRef-Figure-35155131},
and the flow from right to left across the bottom half of Figure \ref{XRef-Figure-35155514}
progressively constructs the probability image.

In Figure \ref{XRef-Figure-35155514} the input image becomes layer
0 of a multilayer network. In layer 0 we extract a pair of adjacent
pixels, and then pass it through a look-up table (or mapping) to yield
a single value which we write into the appropriate pixel location
in layer 1 (in Figure \ref{XRef-Figure-3515426} this corresponds
to transforming $(x_{1},x_{2})$ to become $y_{1}$). We repeat this
operation all over layer 0, to yield a whole array of transformed
values in layer 1. There is an arbitrariness in our choice of the
relative position of the pairs of pixels that we use (e.g. north-south,
or east-west, etc). In our simulations we use a north-south relative
position in the layer 0-to-1 transformation, east-west in the layer
1-to-2 transformation, and alternate these two choices thereafter
as we progress from layer to layer of the network. Note also that
the separation of the pairs of pixels is not the same in each layer.
In Figure \ref{XRef-Figure-35155131} the separation doubles as we
progress from layer to layer, but in Figure \ref{XRef-Figure-35155514}
the separation doubles after every two layers, because we must allow
both the east-west and the north-south orientations to be processed
at all separations (this is a consequence of processing 2-dimensional
data through a 1-dimensional tree-structured network). If we concentrate
only on the topology of the network that results from this prescription
in Figure \ref{XRef-Figure-35155514}, we discover that it is identical
to the topology in Figure \ref{XRef-Figure-35155131}. Thus, the only
difference between these two cases is the way in which we identify
the pixels of the input data array with the layer 0 nodes.

We may use any transformation that we wish in the look-up table. We
have not yet discovered a computationally cheap way of optimising
the network in order to construct a $Q(\text{\boldmath\ensuremath{x}})$
that best approximates the required $P(\text{\boldmath\ensuremath{x}})$.
Instead, we optimise the network in such a way that each layer could
be used to reconstruct approximately the state of the previous layer.
This is not the same optimisation problem, but it is computationally
very cheap, and empirically it leads to useful results for $Q(\text{\boldmath\ensuremath{x}})$.
We choose to train the network as a multilayer topographic mapping,
which we implement in a look-up table after the training schedule
has ended. Typically, the largest number of bits per pixel that we
use is 8, which corresponds to a look-up table with 65536 ($=2^{2\times8}$)
separate addresses, each containing an 8 bit output value.

When we have trained a sufficient number of layers, we may estimate
the clique PDFs in each layer. We simply record these as histograms,
without making any attempt to interpolate or smooth these estimates;
later on we shall mention a number of caveats. This completes the
left-to-right pass in the top half of Figure \ref{XRef-Figure-35155514}.

In order to construct our geometric mean estimate $Q_{\mathrm{gm}}(\text{\boldmath\ensuremath{x}})$
of $P(\text{\boldmath\ensuremath{x}})$, we must combine the estimates
of the clique PDFs. We may obtain the result in Equation \ref{XRef-Equation-35155241}
by appropriately scaling and summing the logarithms of the histograms
(and their marginal histograms) in Figure \ref{XRef-Figure-35155514}.
The method that we use depends on the following rearrangement of Equation
\ref{XRef-Equation-35155241} \begin{equation}
\log Q_{\mathrm{gm}}(\text{\boldmath\ensuremath{x}})=\left(\cdots\,\left(\frac{1}{2}\sum\limits _{k_{n-3}}\log P_{k_{n-3}}^{n-3}+\frac{1}{2}\left(\sum\limits _{k_{n-2}}\log P_{k_{n-2}}^{n-2}+\frac{1}{2}\sum\limits _{k_{n-1}}\log P_{k_{n-1}}^{n-1}\right)\right)\right)\label{XRef-Equation-3516040}\end{equation}

\noindent in which we successively compute the contributions starting
at network layer $n-1$, and then work outwards towards layer 0. First
of all we initialise all of the images in the PDF estimation subsystem
to some constant value (say zero), and then commence at layer $n-1$
(i.e. the righthandmost layer in Figure \ref{XRef-Figure-35155514}).
Using the notation of Figure \ref{XRef-Figure-35154532}, each clique
in the multilayer topographic mapping subsystem contributes a term
of the form $\log P_{\mathrm{out}}(y_{1},y_{2})-\log P_{\mathrm{out}}(y_{1})-\log P_{\mathrm{out}}(y_{2})$,
which we add to the values stored in the two pixels that are located
at the same clique position in the PDF estimation subsystem. In order
to compensate for this double counting, and in order to account for
the $\frac{1}{2}$ factors that appear in Equation \ref{XRef-Equation-3516040},
we scale the logarithmic value by a factor $\frac{1}{4}$ ($=\frac{1}{2}\times\frac{1}{2}$).
We then progress layer by layer towards the left in Figure \ref{XRef-Figure-35155514}.
At each layer we generate its logarithmic contribution as above, but
now we add to this the contribution from the layer on its right, as
shown in Figure \ref{XRef-Figure-35155514} and Equation \ref{XRef-Equation-3516040}.
By cascading the results backwards from layer to layer of the network,
we iteratively construct $\log Q_{\mathrm{gm}}(\text{\boldmath\ensuremath{x}})$
in the form shown in Equation \ref{XRef-Equation-3516040}. Note that
the layer 0 cliques are slightly different, because they contribute
terms of the form $\log P_{\mathrm{in},12}(x_{1},x_{2})$.

When all of these stages are complete, the output image in Figure
\ref{XRef-Figure-35155514} contains pixel values whose sum equals
the required $\log Q_{\mathrm{gm}}(\text{\boldmath\ensuremath{x}})$.
The contribution to $\log Q_{\mathrm{gm}}(\text{\boldmath\ensuremath{x}})$
that is recorded in an output pixel derives from a (rectangular) region
in the input image that surrounds the location of the output pixel,
so the output image can be interpreted as an image of correctly spatially
registered logarithmic probability contributions to $\log Q_{\mathrm{gm}}(\text{\boldmath\ensuremath{x}})$.

In our simulations we investigate how each individual layer of the
multilayer network contributes to $\log Q_{\mathrm{gm}}(\text{\boldmath\ensuremath{x}})$,
so we switch off all except one of the sources of logarithmic probability
in Figure \ref{XRef-Figure-35155514}, which permits only a single
layer of the network to contribute to the construction of the output
image. Because each layer of the network typically is sensitive to
statistical structure in the input image at only one length scale,
the output image then typically reveals contributions to $\log Q_{\mathrm{gm}}(\text{\boldmath\ensuremath{x}})$
at only one length scale.

We should remark in passing that there are many other possible ways
in which Figure \ref{XRef-Figure-35155514} could be configured. Our
results depend on an underlying tree-like structure, which we replicate
to produce the translation invariant network in Figure \ref{XRef-Figure-35155131},
which we then use directly to produce the design in Figure \ref{XRef-Figure-35155514}.
In the case of a non-binary tree we must be careful to produce the
correct generalisation of Figure \ref{XRef-Figure-35155131} and Figure
\ref{XRef-Figure-35155514}, but there are no new difficulties in
principle.

\subsection{Algorithmic Details}

We compensate for some of the effects of non-uniform illumination
of the scene in the input image by adding a grey scale wedge whose
gradient we choose in such a way as to remove the linear component
of the non-uniformity. This improves the assumed translation invariance
of the image statistics. We do not attempt to perform a histogram
equalisation on the input image, because the transformation from network
layer 0 to layer 1 tends to perform this function anyway. In order
not to disrupt the discussion, we review the details of the topographic
mapping training algorithm in the appendix.

We choose to process the image in alternate directions using the following
sequence: north/south, east/west, north/south, east/west, etc. This
sequence leads to the following sequence of rectangular regions of
the input image that influence the value in each pixel in each layer
of the network: $1\times2$, $2\times2$, $2\times4$, $4\times4$,
etc, using (east/west, north/south) coordinates. In all of our experiments
we use a 6-layer network, so the value in each pixel in the final
layer is sensitive to an $8\times8$ region of the input image.

The number of bits per pixel $B$ that we use in each layer of the
network determines the quality of the topographic mappings (the $B$
bit output from a topographic mapping is the index of the winner from
amongst $2^{B}$ competing {}``neurons''). Increasing $B$ improves
the quality of the mapping but increases the training time; we need
to compromise between these two conflicting requirements. In our work
on simple Brodatz texture images we find that choosing $B$ to lie
between 6 and 8 proves to be sufficient. Note that we choose to use
the same number of bits per pixel in each layer of the network. In
general this restriction is not necessary.

It is important to note that for a given value of $B$ there is an
upper limit on the allowed entropy (per unit area) that the input
data should have. A hierarchically connected multilayer topographic
mapping network progressively squeezes the input data through an ever
smaller bottleneck (in fact there are multiple parallel bottlenecks
due to the overlapping tree structure) as we pass through the layers
of the network. There is a upper limit to the number of network layers
beyond which it simply cannot preserve information that is useful
in estimating the joint density of the input data, which limits the
capabilities of our current method.

The choice of the size of the histogram bins is also important. A
property of the multilayer topographic mapping network is that adjacent
histogram bins derive from input vectors that are close to each other
(in the Euclidean sense), so it makes sense to rebin the histogram
by adding together the contents of adjacent bins. We may easily control
the histogram bin size by truncating the low order bits of each pixel
value. If we truncate $b$ low order bits of each pixel value, then
effectively we smooth the histogram over $2^{b}$ adjacent bins (for
each dimension of the histogram). As we smooth the histogram it will
suffer from less noise, but we run the danger of smoothing away significant
structure that might usefully be used to characterise the statistics
of the input image; so we need to make a compromise.

It is most important not to use histogram bins that are too small.
A large number of small histogram bins would record the details of
the statistical fluctuations of the training image (as particular
realisations of a Poisson noise process in each bin), and would act
as a detailed record of the structure in the training image, and thus
be unable to generalise very well. Such histograms would look very
spiky, and in extreme cases there might be counts recorded in only
a few bins, with zeros in all of the remaining bins. If this situation
were to occur, then the training image would have a large $Q_{\mathrm{gm}}(\text{\boldmath\ensuremath{x}})$,
whereas a test image (having the same statistical properties) would
have a small $Q_{\mathrm{gm}}(\text{\boldmath\ensuremath{x}})$. The
cause of this problem is the absence of a significant overlap between
the spikes in the training and test image histograms, which could
be avoided by ensuring that the histogram bins are not too small.
Generally, we find that a little experimentation can be used to determine
a robust histogram binning strategy, so we do not attempt to implement
a more sophisticated technique here.

Finally, we display the contributions to $\log Q_{\mathrm{gm}}(\text{\boldmath\ensuremath{x}})$
as follows. We determine the range of pixel values that occurs in
the image, and we translate and scale this into the range $[0,255]$.
This ensures that the smallest logarithmic probability appears as
black, and the largest logarithmic probability appears as white, and
all other values are linearly scaled onto intermediate levels of grey.
This prescription has its dangers because each image determines its
own special scaling, so one should be careful when comparing two different
images. It can also be adversely affected by pixel value outliers
arising from Poisson noise effects, where an extreme value of a single
pixel could affect the way in which the whole of an image is displayed.
However, we find that the overlapping tree structure of our multilayer
network causes enough averaging together of individual contributions
$Q(\text{\boldmath\ensuremath{x}})$ that the composite result $Q_{\mathrm{gm}}(\text{\boldmath\ensuremath{x}})$
does not suffer from problems due to pixel value outliers.

\section{Application to the Detection of Anomalies in Textures}

\label{XRef-Section-35153635}

In this section we present the results of applying the system shown
in Figure \ref{XRef-Figure-35155514} to four $256\times256$ images
of textures taken from the Brodatz texture album \cite{Brodatz1966}.
In all cases we compensate for uneven illumination by introducing
a grey scale wedge as we explained earlier, we use 8 bits per pixel
for the topographic mappings, we use 6 bits per pixel for histogramming,
and we invert the $[0,255]$ scale to represent the contributions
to $\log Q_{\mathrm{gm}}(\text{\boldmath\ensuremath{x}})$ in such
a way that white pixels indicate a small (rather than a large) contribution
to $\log Q_{\mathrm{gm}}(\text{\boldmath\ensuremath{x}})$. Thus white
pixels in the output image correspond to regions of the input image
whose statistical properties differ markedly from the statistics averaged
over the whole image. We usually call this representation of the contributions
to $\log Q_{\mathrm{gm}}(\text{\boldmath\ensuremath{x}})$ an {}``anomaly
image''.

We do not present these results as necessarily being an efficient
way of detecting texture anomalies. Rather, we merely apply our novel
method of estimating PDFs, as expressed in Equation \ref{XRef-Equation-35155241}
and in Figure \ref{XRef-Figure-35155514}, to the particular problem
of texture analysis, because this is an effective way of demonstrating
some of the more interesting properties of $\log Q_{\mathrm{gm}}(\text{\boldmath\ensuremath{x}})$.

\subsection{Texture 1}

\begin{figure}[H]
\begin{centering}
\includegraphics[width=7cm]{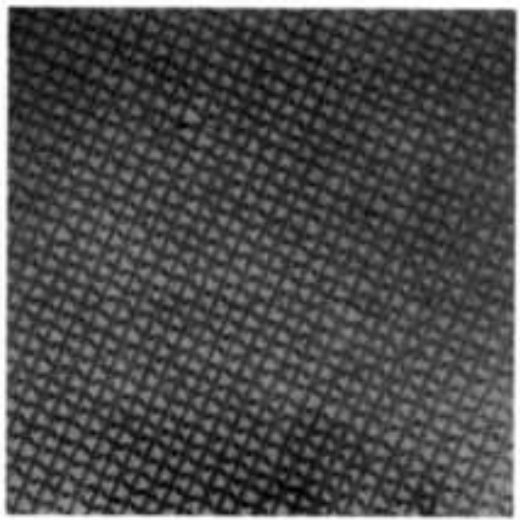}
\par\end{centering}

\caption{256$\times$256 image of Brodatz image 1.}

\label{XRef-Figure-3516836}
\end{figure}

In Figure \ref{XRef-Figure-3516836} we show the first Brodatz texture
image that we use in our experiments. The image is slightly unevenly
illuminated and has a fairly low contrast, but nevertheless its statistical
properties are almost translation invariant. %

\begin{figure}[H]

\begin{centering}
\includegraphics[width=7cm]{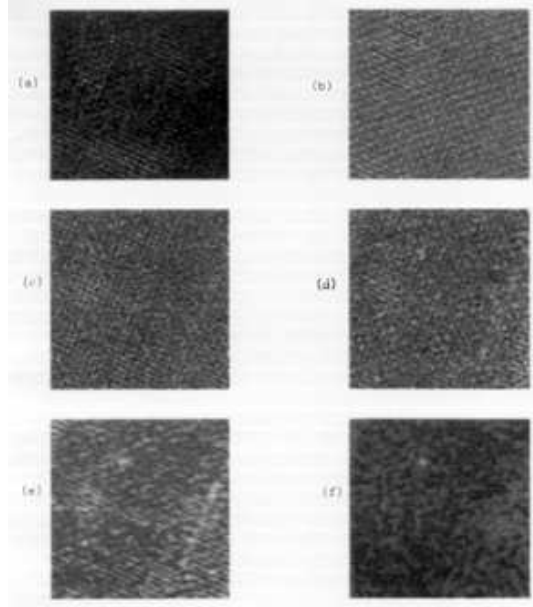}
\par\end{centering}

\caption{256$\times$256 anomaly images of Brodatz image 1.}

\label{XRef-Figure-3516937}
\end{figure}

In Figure \ref{XRef-Figure-3516937} we show the anomaly images that
derive from Figure \ref{XRef-Figure-3516836}. Note how the anomaly
images become smoother as we progress from Figure \ref{XRef-Figure-3516937}a
to\ \ Figure \ref{XRef-Figure-3516937}f, due to the increasing
amount of averaging that occurs amongst the overlapping trees in the
network. %

Figure \ref{XRef-Figure-3516937}e and Figure \ref{XRef-Figure-3516937}f
reveal a highly localised anomaly in the original image. Figure \ref{XRef-Figure-3516937}f
corresponds to a length scale of $8\times8$ pixels, which is the
approximate size of the fault that is about $\frac{1}{4}$ of the
way down and slightly to the left of centre of Figure \ref{XRef-Figure-3516836}.
The fault does not show up clearly on the other figures in Figure
\ref{XRef-Figure-3516937} because their characteristic length scales
are either too short or too long to be sensitive to the fault.

From Figure \ref{XRef-Figure-3516937} we conclude that ACE can easily
pick out localised faults in highly ordered textures.

\subsection{Texture 2}

\begin{figure}[H]

\begin{centering}
\includegraphics[width=7cm]{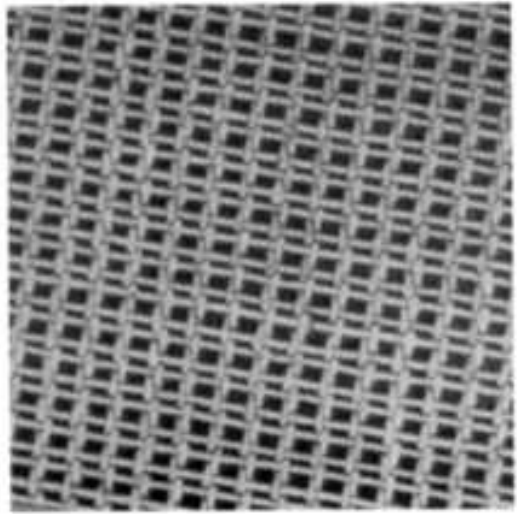}
\par\end{centering}

\caption{256$\times$256 image of Brodatz image 2.}

\label{XRef-Figure-35161234}
\end{figure}

In Figure \ref{XRef-Figure-35161234} we show the second Brodatz texture
image that we use in our experiments. The image has a high contrast
and translation invariant statistical properties. %
\begin{figure}[H]

\begin{centering}
\includegraphics[width=7cm]{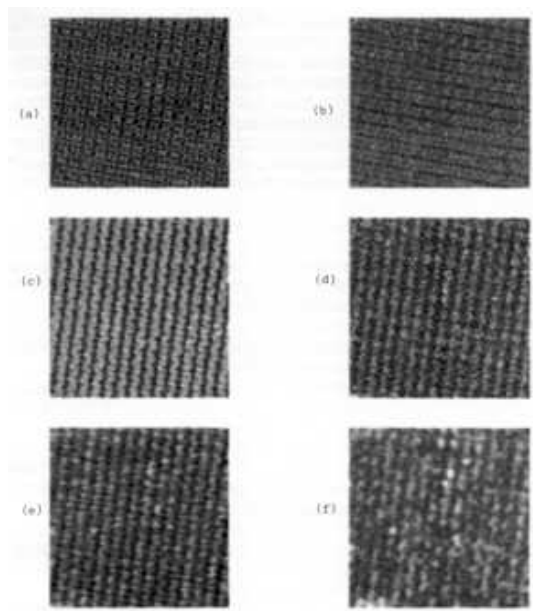}
\par\end{centering}

\caption{256$\times$256 anomaly images of Brodatz image 2.}

\label{XRef-Figure-35161531}
\end{figure}

In Figure \ref{XRef-Figure-35161531} we show the anomaly images that
derive from Figure \ref{XRef-Figure-35161234}. The most interesting
anomaly image is Figure \ref{XRef-Figure-35161531}f which shows several
localised anomalies. About halfway down and to the left of centre
of the image is an anomaly that corresponds to a dark spot on the
thread in Figure \ref{XRef-Figure-35161234}. The brightest of the
anomalies in the cluster just above the centre of the image corresponds
to what appears to be a slightly torn thread in Figure \ref{XRef-Figure-35161234}.
The other anomalies in this cluster are weaker, and correspond to
slight distortions of the threads. There is another anomaly just below
and to the right of the centre of Figure \ref{XRef-Figure-35161531}f,
which corresponds to what appears to be another slightly torn thread
in Figure \ref{XRef-Figure-35161234}. These anomalies all occur at,
or around, a length scale of $8\times8$ pixels. Several of the anomaly
images show an anomaly in the bottom left hand corner of the image,
which corresponds to a small uniform patch of fabric in Figure \ref{XRef-Figure-35161234}.

The results in Figure \ref{XRef-Figure-35161531} corroborate the
evidence in Figure \ref{XRef-Figure-3516937} that we can train ACE
to pick out localised anomalies in highly structured textures. This
type of texture could be analysed much more simply by model-based
techniques that took advantage of their near-periodicity. However,
that does not detract from the fact that, by making use of its adaptability,
ACE succeeds in modelling these textures without prior knowledge of
their near-periodicity. We seek a general purpose approach to density
estimation; not a toolkit of different (usually model-based) techniques,
each tuned to its own type of problem.

\subsection{Texture 3}

\begin{figure}[H]

\begin{centering}
\includegraphics[width=7cm]{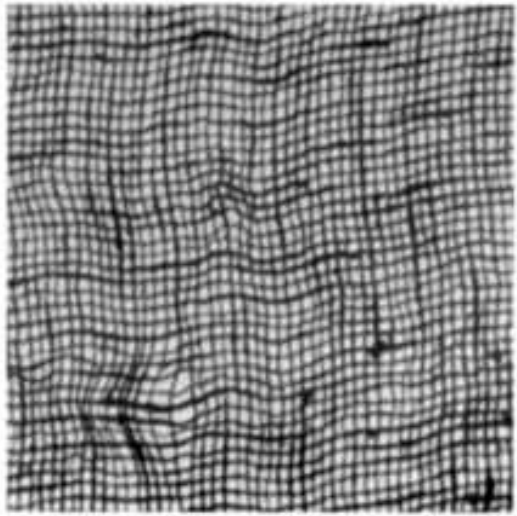}
\par\end{centering}

\caption{256$\times$256 image of Brodatz image 3.}

\label{XRef-Figure-35162040}
\end{figure}

In Figure \ref{XRef-Figure-35162040} we show the third Brodatz texture
image that use in our experiments. The image has a very high contrast
and statistical properties that are almost translation invariant.
However the density of anomalies is much higher than in either Figure
\ref{XRef-Figure-3516836} or Figure \ref{XRef-Figure-35161234}.
\begin{figure}[H]

\begin{centering}
\includegraphics[width=7cm]{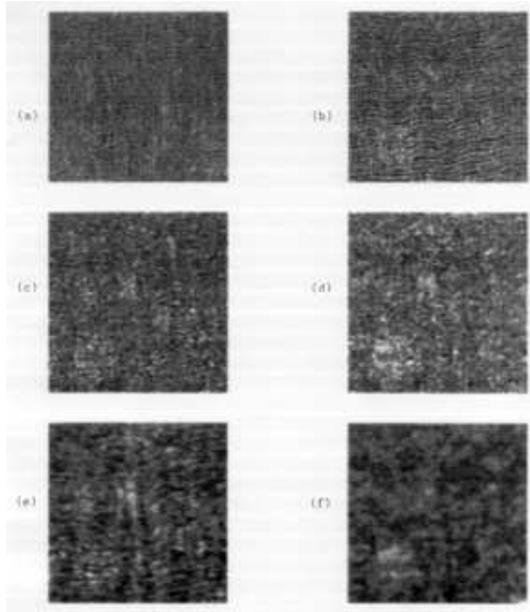}
\par\end{centering}

\caption{256$\times$256 anomaly images of Brodatz image 3.}

\label{XRef-Figure-35162120}
\end{figure}

In Figure \ref{XRef-Figure-35162120} we show the anomaly images that
derive from Figure \ref{XRef-Figure-35162040}. At the lower left
hand corner of Figure \ref{XRef-Figure-35162120}f there is a large
anomaly that corresponds to a region of Figure \ref{XRef-Figure-35162040}
that is distorted to the left. Figure \ref{XRef-Figure-35162120}f
is sensitive to a length scale of $8\times8$ pixels, so it does not
respond to this leftward distortion (which occurs on a length scale
of around $32\times32$ pixels), rather it responds to localised variations
in the separations of the threads.

There are numerous other anomalies in Figure \ref{XRef-Figure-35162040};
some are detected in Figure \ref{XRef-Figure-35162120}, and some
are not. The ability of ACE to pick out anomalies degrades as the
density of anomalies increases. This is because the anomalies themselves
are part of the statistical properties that are extracted by ACE from
the training image, and if a particular type of anomaly occurs often
enough in the image then it is no longer deemed to be an anomaly.
In extreme cases there is also the possibility that the entropy (per
unit area) of the input image can saturate ACE and thus degrade its
performance, as we discussed earlier.

\subsection{Texture 4}

In this section we present a slightly different type of experiment
in which we train ACE on one image and test ACE on another image.
To create the two images we start with a single $256\times256$ image
of a Brodatz texture, which we divide into a left half and a right
half. We then use the left half to construct a training image, and
the right half to construct a test image. Note that in constructing
these images we scrupulously avoid the possibility that the training
and test images contain elements deriving from a common source, although
there are some small residual correlations between the two images
along their common edge. %
\begin{figure}[H]

\begin{centering}
\includegraphics[width=7cm]{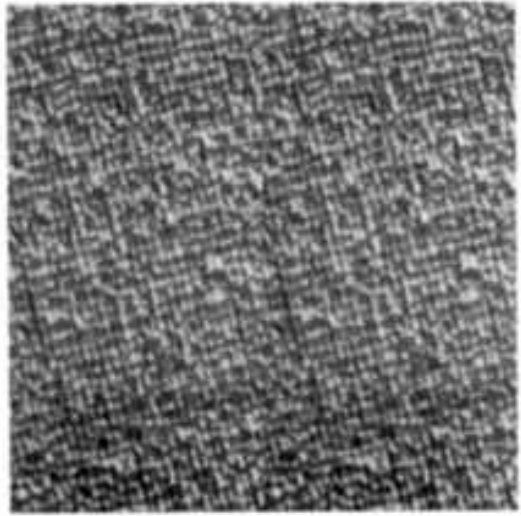}
\par\end{centering}

\caption{256$\times$256 image of a Brodatz image of a carpet for training.}

\label{XRef-Figure-35162456}
\end{figure}

In Figure \ref{XRef-Figure-35162456} we show the training image which
is a montage of two copies of the left hand half of a Brodatz texture
image. Note that we use square, rather than rectangular, images because
our software is restricted to processing this type of image. %
\begin{figure}[H]

\begin{centering}
\includegraphics[width=7cm]{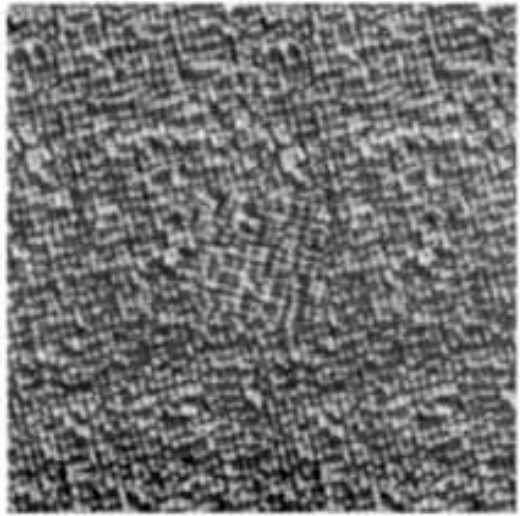}
\par\end{centering}

\caption{256$\times$256 image of a Brodatz image of a carpet for testing.}

\label{XRef-Figure-35162516}
\end{figure}

In Figure \ref{XRef-Figure-35162516} we show the test image which
is a montage of two copies of the right hand half of a Brodatz texture
image, and superimposed on that is a $64\times64$ patch which we
generated by flipping the rows and columns of a copy of the top left
hand corner of this image. This patch is a hand-crafted anomaly. %
\begin{figure}[H]

\begin{centering}
\includegraphics[width=7cm]{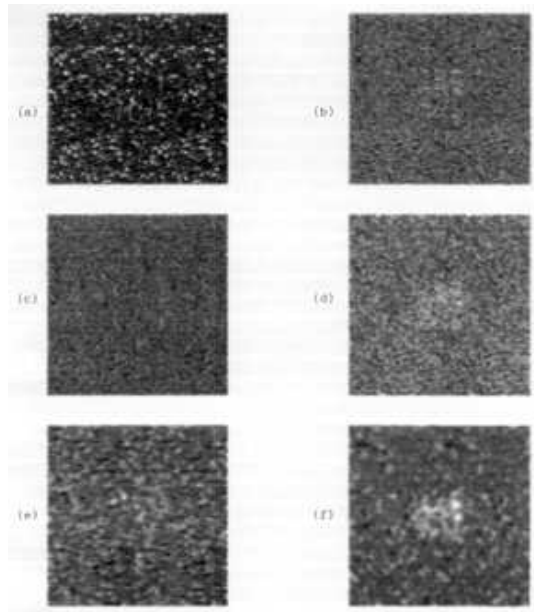}
\par\end{centering}

\caption{256$\times$256 anomaly images of a Brodatz image of a carpet.}

\label{XRef-Figure-35162534}
\end{figure}

In Figure \ref{XRef-Figure-35162534} we show the anomaly images that
derive from Figure \ref{XRef-Figure-35162516} after we train on Figure
\ref{XRef-Figure-35162456}. Figure \ref{XRef-Figure-35162534}f shows
the strongest response to the anomalous patch in the centre of the
image, corresponding to anomaly detection on a length scale of $8\times8$
pixels.

\section{Conclusions}

We present a novel method of density estimation in high-dimensional
spaces, such as images. In Bayesian data processing there is a pressing
need for a flexible way of constructing such estimates, because the
basic objects that we manipulate in Bayesian analysis are joint PDFs,
which we must somehow construct in the first place. We call the hierarchical
network structure that emerges from our analysis an {}``Adaptive
Cluster Expansion'', or ACE for short.

ACE is computationally very cheap: we can train a multilayer topographic
mapping network to estimate the joint PDF of its input data at the
rate of 1 network layer every 2.3 second (on a VAXstation 3100, and
assuming 6 bits per pixel), where each layer analyses one length scale
(power of 2) in the input data. We find, in our experiments with Brodatz
textures, that 6 network layers allows the detection of statistical
anomalies in the textures. This result is not universal, because it
must depend strongly on the scale at which the anomalous statistical
structure in the data is to be found. Although we demonstrate ACE
only in a texture anomaly detection rôle, its scope is far greater
than this. ACE is a general purpose, and computationally cheap, network
for estimating densities in high-dimensional spaces.

For completeness, we should mention that the performance of ACE in
its current form has two fundamental limitations. Firstly, we assume
that the network connectivity is fixed, and that its functionality
is determined by a training algorithm. This restricts the possible
statistical properties of the input data that could be estimated.
Secondly, ACE is based upon a hierarchically connected multilayer
topographic mapping network, which progressively squeezes the input
data through an ever smaller bottleneck as we pass through the layers
of the network. There is a upper limit to the number of network layers
beyond which ACE simply cannot preserve information that is useful
in estimating the statistics of the input data. For instance, the
statistics of an extremely noisy image of a texture can not be successfully
estimated by ACE, because the noise entropy would saturate ACE before
the statistics of the underlying texture could be investigated. This
problem can be solved by introducing explicit noise models into ACE,
which we shall report elsewhere.

\appendix

\section{Appendix}

The standard topographic mapping training procedure in \cite{Kohonen1984}
is a rather inefficient algorithm. In \cite{Luttrell1989e} we present
in detail an efficient training procedure for topographic mappings,
and explain how to use it to train multilayer topographic mappings.

For convenience, we introduce some notation.

\ $\text{\boldmath\ensuremath{x}}$ = input vector

\ $y$ = index of the winning {}``neuron''

\ $\text{\boldmath\ensuremath{x}}(y)$ = reference vector associated
with $y$

\ $\pi(y^{\prime}-y)$ = topographic neighbourhood function (normalised
to unit total mass)

\ $\varepsilon$ = update parameter used during training

\ $N$ = number of reference vectors

\subsection{Standard Topographic Mapping Training Algorithm}

The standard topographic mapping training procedure is essentially
as follows \cite{Kohonen1984}:
\begin{enumerate}
\item Select a training vector $\text{\boldmath\ensuremath{x}}$ at random
from the training set.\label{Step:1}
\item Map $\text{\boldmath\ensuremath{x}}$ to $y$ by using a nearest neighbour
prescription applied to the distance of $\text{\boldmath\ensuremath{x}}$
from each of the current set of reference vectors.
\item For all $y^{\prime}$, move the reference vector $\text{\boldmath\ensuremath{x}}(y^{\prime})$
directly towards the input vector $\text{\boldmath\ensuremath{x}}$
by a distance $\varepsilon\,\pi(y^{\prime}-y)\,\|\text{\boldmath\ensuremath{x}}-\text{\boldmath\ensuremath{x}}(y^{\prime})\|$.
\item Go to step 1.
\end{enumerate}
Repeat this loop as often as is required to ensure convergence of
the reference vectors.

The standard training method specifies that $\pi(y^{\prime}-y)$ should
be an even unimodal function whose width should be gradually decreased
as training progresses. This allows coarse-grained organisation of
the reference vectors to occur, followed progressively by ever more
fine-grained organisation, until finally the algorithm converges to
an optimum set of reference vectors. In a similar vein, the relative
size of the update step $\varepsilon$ should also be steadily decreased
as training progresses.

\subsection{Modified Topographic Mapping Training Algorithm}

In our own modification \cite{Luttrell1989e} of the standard topographic
mapping training we replace a shrinking $\pi(y^{\prime}-y)$ function
acting on a fixed number of reference vectors, by a fixed $\pi(y^{\prime}-y)$
function acting on an increasing number of reference vectors. There
are many minor variations on this theme, but we find that it is sufficient
to define \[
\pi(y^{\prime}-y)=\left\{ \begin{array}{lcl}
\varepsilon &  & y^{\prime}=y\\
\varepsilon^{\prime} &  & \left|y^{\prime}-y\right|=1\\
0 &  & \left|y^{\prime}-y\right|>1\end{array}\right.\]

\noindent where we absorb the $\varepsilon$ into the definition of
$\pi(y^{\prime}-y)$. We increase the number of reference vectors
in a binary sequence (i.e. $N=2,4,8,16,32,\cdots$), and we initialise
each generation of reference vectors by interpolation from the previous
generation. We find that the following parameter values yield adequate
convergence: $\varepsilon=0.1$, $\varepsilon^{\prime}=0.05$, and
we perform $20N$ training updates before doubling the value of $N$,
as above. We initialise the $N=2$ pair of reference vectors as a
random pair of vectors chosen from the training set.

In numerous experiments, we find that this modified form of the topographic
mapping training algorithm converges much more rapidly than the standard
method. Furthermore, the binary sequence of $N$ values lends itself
well to implementing the trained topographic mapping using a look-up
table.

\end{document}